\documentclass[11pt,a4paper]{article}
\usepackage[hyperref]{naaclhlt2019}
\usepackage{times}
\usepackage{latexsym}
\usepackage{booktabs}
\usepackage{graphicx}
\usepackage{array}
\newcolumntype{C}[1]{>{\centering\let\newline\\\arraybackslash\hspace{0pt}}m{#1}}
\usepackage{xspace}
\usepackage{soul}
\usepackage{tikz}\usetikzlibrary{positioning,arrows,shapes}
\usepackage{caption}
\usepackage{subcaption}

\usepackage{xcolor}
\usepackage{color}
\usepackage[framemethod=tikz]{mdframed}

\newcommand{\mparagraph}[1]{\noindent\textbf{{#1}}.}
\newcommand{\mparagraphnp}[1]{\noindent\textbf{{#1}}}

\newtoggle{hideOurComments}
\togglefalse{hideOurComments} \toggletrue{hideOurComments}

\newcommand{\pdfcomment}[2]{\textcolor{#1}{#2}} \newcommand{\llee}[1]{\pdfcomment{magenta}{#1}}
\newcommand{\jack}[1]{\pdfcomment{blue}{#1}}

\iftoggle{hideOurComments}{
\renewcommand{\pdfcomment}[2]{}
}

\newcommand{\subreddit}[1]{{\sf #1}}  \newcommand{\pupv}{percent-upvoted\xspace} 

\definecolor{lightgray}{rgb}{.3,.3,.3}

\aclfinalcopy \def\aclpaperid{1616}

\title{Something's Brewing!\\Early Prediction of Controversy-causing Posts
from Discussion Features
}

\author{
  Jack Hessel \\
  Cornell University \\
  {\tt jhessel@cs.cornell.edu} \\\And
  Lillian Lee \\
  Cornell University \\
  {\tt llee@cs.cornell.edu}
}
\date{}

\begin{document}

\maketitle

\begin{abstract}
                        Controversial posts are those that split the preferences of a
  community, receiving both significant positive and significant negative feedback.
  Our inclusion of the word ``community'' here is deliberate: what is
  controversial to some audiences may not be so to others.
  Using data from several different communities on \texttt{reddit.com},
        we predict the ultimate
  controversiality of posts, leveraging
      features drawn from both the textual content and the tree structure of the
  early comments that initiate the discussion.
              We find that even when only a handful of comments are available,
  e.g., the first 5 comments made within 15 minutes of the original post,
  discussion features often add predictive capacity to strong
  content-and-rate only baselines.
  Additional experiments on domain transfer suggest that conversation-structure
  features often generalize to other communities better than conversation-content
  features do.

\end{abstract}

\section{Introduction}
\label{sec:intro}

Controversial content --- that which attracts both positive and
negative feedback --- is not necessarily a bad thing; for instance, bringing
up a point that warrants spirited debate can improve community health.\footnote{\citet{Coser:1956,jehn_multimethod_1995,DeDreu+Weingart:2003} discuss how
disagreement interacts with group makeup, group-task type,
and outcome. \citet{chen2013and} demonstrate a non-linear relationship between controversy and amount of subsequent discussion. }
But regardless of the nature of the controversy,
detecting potentially controversial content can be useful
for both community members and community moderators.  Ordinary users,
and in particular new users, might appreciate being warned that they
need to add more nuance or qualification to their earlier
posts.\footnote{We set aside the issue of {\em trolls}
whose intent is solely to divide a community.
}
Moderators could be alerted
that the discussion ensuing from some content might need
monitoring.  Alternately, they
could draw community attention to issues
possibly needing resolution:
indeed, some sites
already provide explicit sorting by
controversy.

We consider the controversiality of a piece of content in the context
of the community in which it is shared, because what is controversial
to some audiences may not be so to others \citep{chen2013and,jang_modeling_2017,basile2017predicting}.
For example, we identify ``break up'' as a controversial concept in
the \subreddit{relationships} subreddit
(a subreddit is a subcommunity hosted on the Reddit discussion site),
but the same topic
is associated with a \emph{lack} of controversy in the
\subreddit{AskWomen} subreddit
(where questions are posed for women to answer).
Similarly, topics that are controversial in one community may
simply not be discussed in another: our analysis identifies ``crossfit'', a
type of workout, as one of the most controversial concepts
in the subreddit \subreddit{Fitness}.

However, while controversial topics may be community-specific,
community moderators still may not be able to  determine a priori which
posts will attract controversy. Many factors cannot be known
ahead of time, e.g., a fixed set of topics may not be dynamic enough
to handle a sudden current event, or the specific set of users that
happen to be online at a given time may
react in unpredictable
ways.
Indeed, experiments have shown that, to a certain extent,
the influence of early opinions on subsequent opinion dynamics
can override the influence of an item's actual content
\cite{salganik2006experimental,Wu:2008:POF:1504941.1504989,muchnik2013socialinfluencebias,weninger2015random}.

\newcommand{\refsinfig}[1]{{\scriptsize #1}}

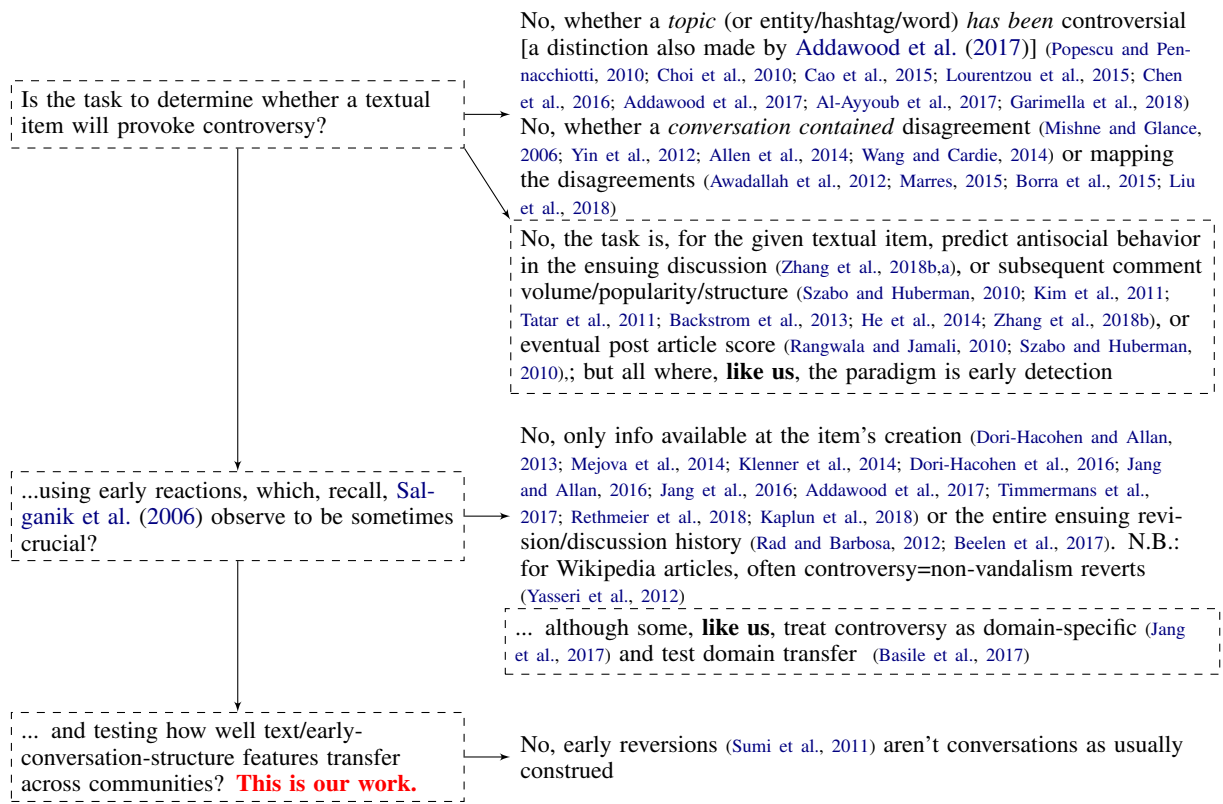
\begin{figure*}[th]
  \centering
  
    \begin{tikzpicture}[auto,
        block_q/.style={rectangle, draw=black, dashed, text width=2.5in, font=\small},
        block_cites/.style={rectangle, text width=3.6in, text ragged, font=\small}, block_cites_add/.style={rectangle,draw=black, dashed, text width=3.6in, text ragged, font=\small},
        line/.style={draw, -latex'}]

    \node (root) [block_q, text width=5.7cm] {Is the task to determine whether a textual item will provoke  controversy?};
    \node (topic) [block_cites,right=.6cm of root, text width=9cm] {No, whether a \emph{topic} (or  entity/hashtag/word)  \emph{has been}
    controversial [a distinction also made by \citet{addawood2017telling}]
    \refsinfig{\cite{popescu2010detecting,choi2010identifying,cao2015socialhelix,lourentzou2015hotspots,chen_wordforce_C16-2057,addawood2017telling,al2017studying,garimella2018quantifying}} \\
    No, whether a \emph{conversation} \emph{contained} disagreement \refsinfig{\cite{mishne2006leave,yin2012unifying,allen2014detecting,wang2014dispute}} or mapping the disagreements
    \refsinfig{\cite{awadallah_harmony_2012,marres_why_2015,borra_societal_2015,liu_consensus:_2018}}
    };     \node (early_other) [block_cites_add, below=-.1cm of topic, text width=9cm] {No, the task is, for the given textual item, predict antisocial behavior in the ensuing discussion \refsinfig{\cite{zhang2018characterizing,Zhang+al:18a}}, or  subsequent comment volume/popularity/structure \refsinfig{\cite{szabo2010predicting,kim2011predicting,tatar2011predicting,Backstrom:ProceedingsOfWsdm:2013,he2014predicting,zhang2018characterizing}}, or eventual post article score \refsinfig{\cite{rangwala2010defining,szabo2010predicting},}; but all where, {\bf like us}, the paradigm is early detection};

    \node (posttime_only) [block_cites,below=.3cm of early_other,text width=9cm] {No, only info available at the item's creation
    \refsinfig{\cite{dori2013detecting,mejova2014controversy,klenner2014verb,Dori-Hacohen:2016:collective,jang2016improving,jang2016probabilistic,addawood2017telling,timmermans2017controcurator,rethmeier2018learning,kaplun_controversy_2018}} or the entire ensuing revision/discussion history \refsinfig{\citep{rad_identifying_2012,beelen_detecting_2017}}.
    N.B.: for Wikipedia articles, often controversy=non-vandalism reverts \refsinfig{\cite{yasseri_dynamics_2012}}};
    \node (posttime_and_comm) [block_cites_add, below=-.1cm of posttime_only]
   {... although some, {\bf like us},  treat controversy as domain-specific \refsinfig{\cite{jang_modeling_2017}} and test domain transfer \refsinfig{ \cite{basile2017predicting}}};

  \node (early_q) [block_q,left= .6cm of posttime_only, text width=5.7cm] {...using early reactions, which, recall, \citet{salganik2006experimental} observe to be sometimes crucial?};

   \node (comm_q) [block_q,below=2cm of early_q, text width=5.7cm] {... and testing how well text/early-conversation-structure features transfer across communities? \textcolor{red}{{\bf This is our work.}}};
   \node (early_contr) [block_cites, right= .6cm of comm_q] {No, early reversions \refsinfig{\cite{sumi_characterization_2011}} aren't conversations as usually construed };

    \begin{scope}[every path/.style=line]
         \path (root) -- (topic);
         \path (root) -- (early_q);
         \path (early_q) -- (posttime_only);
         \path (early_q) -- (comm_q);
         \path (comm_q) -- (early_contr);
         \path (root.south east) -- (early_other.north west);
    \end{scope}
        \end{tikzpicture}
        \caption{How our research relates to prior work.}
    \label{fig:related_laid_out}
\end{figure*}

\label{sec:related}

\llee{mention that Garimella did not find content particularly effective?  Say later that popular + polarizing happens, unlike amendola2015evolving-- although i ended up cutting that reference.}

Hence, we propose an early-detection approach that uses not just the content of the
initiating post, but also the content and structure of the initial responding comments.
In doing so, we unite
streams of heretofore mostly disjoint
research programs: see Figure \ref{fig:related_laid_out}.
Working with over 15,000 discussion trees across six subreddits,
we find that
incorporating structural and textual features of budding comment trees
improves predictive performance relatively quickly; for example, in
one of the communities we consider,
adding features taken from just the first 15
minutes of discussion
significantly increases prediction performance,
even though the average thread
only contains 4 comments by that time
({\raise.17ex\hbox{$\scriptstyle\sim$}}4\% of all eventual comments).

Additionally, we
study feature transferability across domains (in our case, communities),
training on one subreddit and testing on another. While text features of
comments carry the greatest predictive capacity in-domain, we find
that discussion-tree and -rate features are less brittle, transferring
better between communities.

 Our results not only suggest the potential
  usefulness
  of granting controversy-prediction algorithms a small observation
  window to gauge community feedback, but also demonstrate the
  utility of our expressive feature set for early discussions.

\section{Datasets}
\label{sec:dataset}

Given our interest in community-specific controversiality, we draw
data from \texttt{reddit.com},
which hosts several
thousand discussion subcommunities (subreddits) covering a variety of
interests.
Our dataset, which attempts to cover
all public posts and comments
from Reddit's inception in 2007 until Feb. 2014, is derived from a
combination of Jason Baumgartner's posts and comments
sets and our own scraping
efforts
 to fill in dataset gaps. The result is a mostly-complete set
 of posts alongside associated comment trees.\footnote{
Data hosted at \url{pushshift.io}, an open data initiative.
Scraping was performed using Reddit's API or \texttt{github.com/pushshift/api}.
   Roughly 10\% of
comments and 20\% of posts are deleted by users and/or moderators;
also, authorship information is not available for many posts due to
deletion of accounts.}
We focus on six text-based\footnote{
We ignore subreddits devoted to
image sharing.} subreddits ranging over a variety of styles and topics:
two Q\&A subreddits:
\subreddit{AskMen} (AM) and \subreddit{AskWomen}
(AW); a
special-interest community, \subreddit{Fitness} (FT); and three advice communities: \subreddit{LifeProTips} (LT), \subreddit{personalfinance} (PF), and
\subreddit{relationships} (RL).
Each comprises tens of thousands of posts and
hundreds of thousands to millions of comments.

\begin{figure*}
  \centering
  \begin{minipage}{.66\textwidth}
    \centering
    \includegraphics[width=1.0\linewidth]{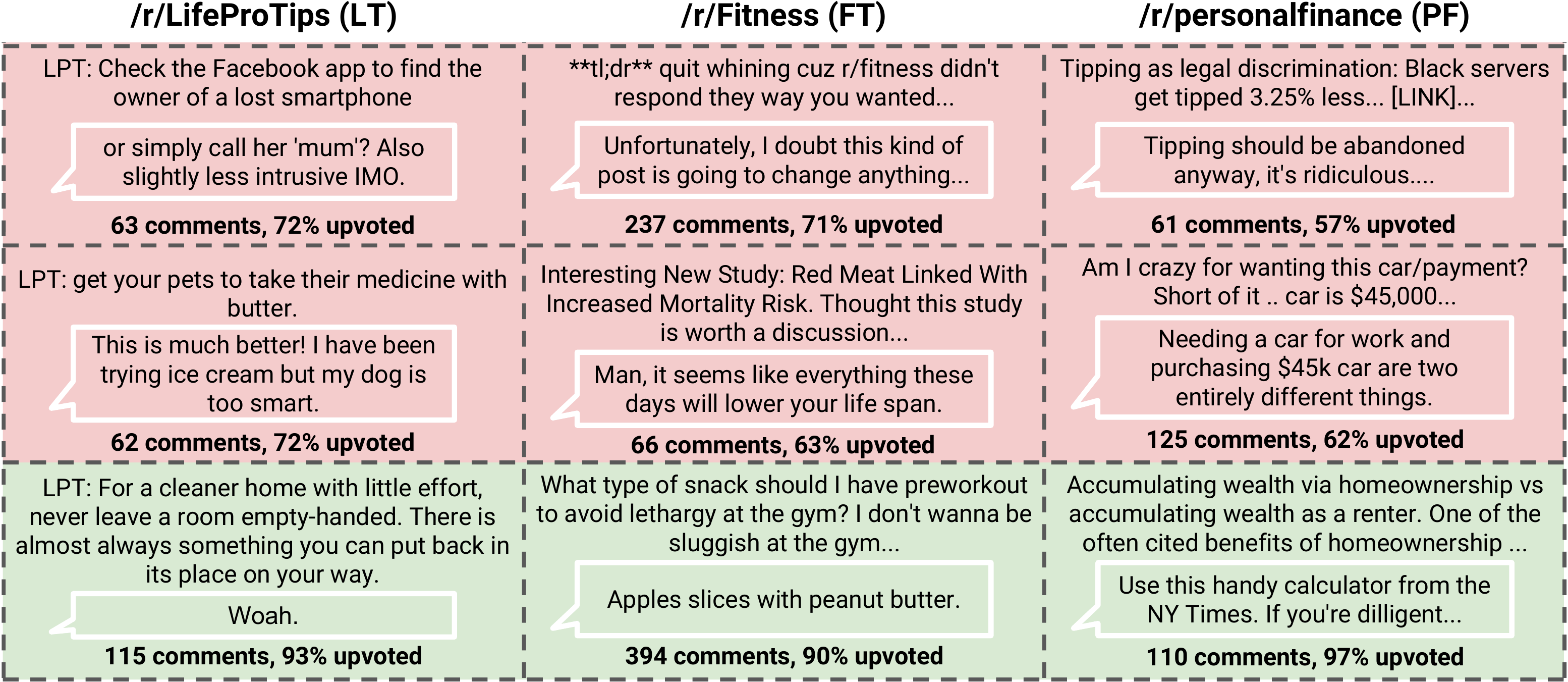}
    \captionof{figure}{Examples of two controversial and one
      non-controversial post from three communities. Also shown are the text of the
      first reply, the number of comments the post received, and its
      \pupv.}
    \label{fig:examples}
  \end{minipage}%
  \hspace{0.50mm}
  \begin{minipage}{.33\textwidth}
    \centering
    \begin{subfigure}{.45\linewidth}
      \centering
      \includegraphics[width=.95\linewidth]{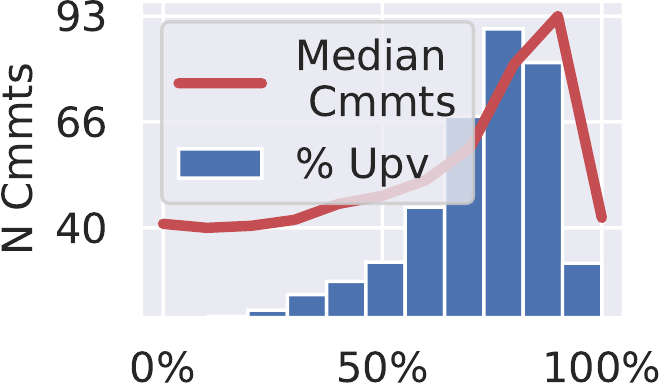}
      \caption{AM}
      \label{fig:AM_outcome}
    \end{subfigure}    \begin{subfigure}{.45\linewidth}
      \centering
      \includegraphics[width=.95\linewidth]{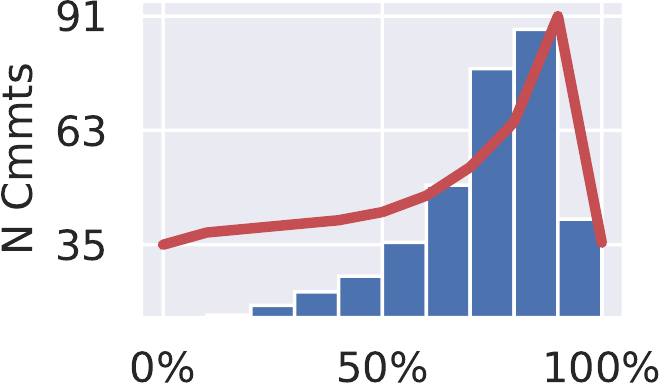}
      \caption{AW}
      \label{fig:AW_outcome}
    \end{subfigure}

    \centering
    \begin{subfigure}{.45\linewidth}
      \centering
      \includegraphics[width=.95\linewidth]{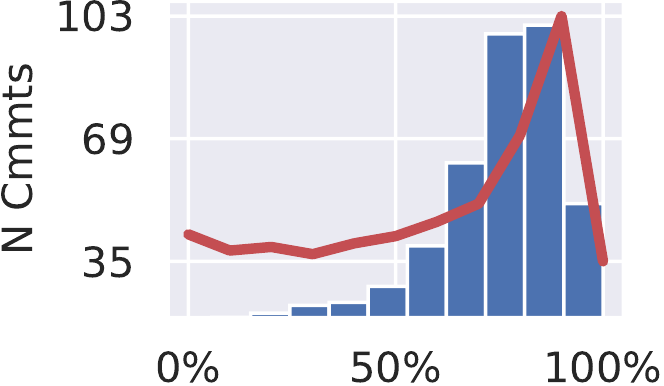}
      \caption{FT}
      \label{fig:FT_outcome}
    \end{subfigure}    \begin{subfigure}{.45\linewidth}
      \centering
      \includegraphics[width=.95\linewidth]{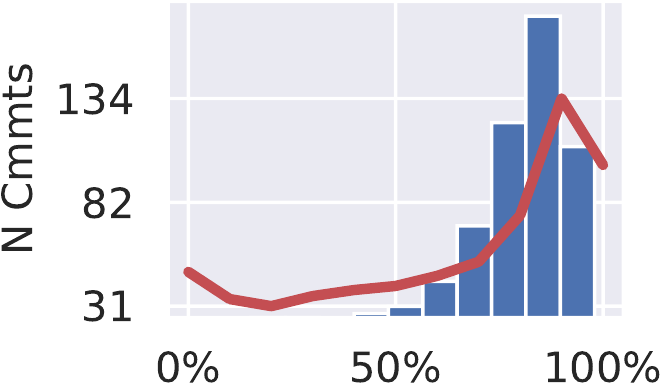}
      \caption{LT}
      \label{fig:LT_outcome}
    \end{subfigure}

    \centering
    \begin{subfigure}{.45\linewidth}
      \centering
      \includegraphics[width=.95\linewidth]{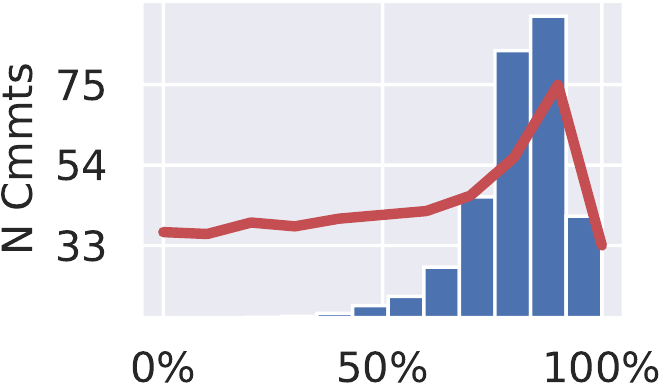}
      \caption{PF}
      \label{fig:PF_outcome}
    \end{subfigure}    \begin{subfigure}{.45\linewidth}
      \centering
      \includegraphics[width=.95\linewidth]{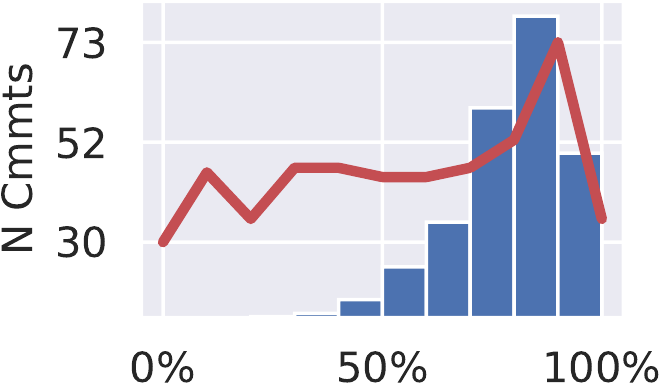}
      \caption{RL}
      \label{fig:RL_outcome}
    \end{subfigure}
    \caption{For each community, a histogram of \pupv and
      the median number of comments per bin.}
    \label{fig:outcomes}
  \end{minipage}
\end{figure*}

In Reddit (similarly to other sites allowing  explicit negative feedback, such as YouTube, imgur, 9gag, etc.), users can give posts {\em upvotes}, increasing a
post's score, or {\em downvotes}, decreasing it.\footnote{Vote timestamps are not publicly available.}  While the semantics of up/down votes may
vary based on community (and, indeed, each  user may have
their own views on what content should be upvoted and what
downvoted), in aggregate, posts that split community reaction fundamentally
differ from those that produce agreement. Thus, in principle, posts that have
unambiguously received both many upvotes \emph{and} many downvotes
should be
deemed the most controversial.

\mparagraph{Percent Upvoted on Reddit} We quantify the relative
proportion of upvotes and downvotes on a post using \emph{\pupv}, a measure
provided by Reddit that gives an estimate of the percent of all votes
on a post that are upvotes.  In practice, exact values of \pupv are
not directly available; the site adds ``vote fuzzing'' to fight
vote manipulation.\footnote{Prior to Dec. 2016,
  vote information was fuzzed according to a different algorithm;
  however, vote statistics for all posts were recomputed according to
  a new algorithm that, according to a reddit moderator, can
  ``actually be trusted;'' \url{https://goo.gl/yHWeJp}} To begin with,
we first discard posts with fewer than 30 comments.\footnote{The
  intent is to only consider posts receiving enough community
  attention  for us to reliably  compare upvote counts with
  downvotes.  We use number of comments as a proxy for aggregate
  attention because Reddit does not surface the true number of votes.}
Then, we query for the noisy \pupv from each post ten times using the
Reddit API, and take a mean to produce a final estimate.

\mparagraph{Post Outcomes}
To better understand the
interplay between
upvotes and downvotes, we first explore
the outcomes for posts both in terms of \pupv and the number of comments;
doing so on a per-community basis has the potential to surface
any subreddit-specific effects. In
addition, we compute the median number of comments for posts falling
into each bin of the histogram.
The resulting plots are
given in Figure~\ref{fig:outcomes}.

In general, posts receive mostly positive feedback in aggregate, though the mean
\pupv varies between communities (Table~\ref{tab:basic_stats}). There is also a positive
correlation between a post's \pupv and the number of
comments it receives.
This relationship is unsurprising, given
that Reddit displays higher rated posts to more users.

A null hypothesis, which we compare to empirically
in our prediction experiments, is that popularity and \pupv simply
carry the same information. However, we have reason to doubt this null
hypothesis, as quite a few posts receive significant attention
despite having a low \pupv (Figure~\ref{fig:examples}).

\mparagraph{Assigning Controversy Labels To Posts} We assign binary
controversy labels (i.e., relatively controversial vs.  relatively
non-controversial) to posts according to the following process: first,
we discard posts where the observed variability across 10 API queries
for \pupv exceeds 5\%; in these cases, we assume that there are too
few total votes for a stable estimate. Next, we discard posts where
neither the observed upvote ratio nor the observed score\footnote{A
  score is the (noised) upvotes minus the downvotes.} vary at all; in these cases, we cannot be sure that the
upvote ratio is insensitive to the vote fuzzing function.\footnote{We
  validate our filtration process in a later section by directly
  comparing to Reddit's rank-by-controversy function.} Finally, we
sort each community's surviving posts by upvote percentage, and
discard the small number of posts with \pupv below
50\%.\footnote{Reddit provides less information for posts with more
  upvotes than downvotes.}
    The top quartile of posts according to this ranking (i.e., posts with
mostly only upvotes) are labeled  ``non-controversial.'' The bottom quartile of posts,
where the number of downvotes cannot exceed but may approach the
number of upvotes, are labeled as ``controversial.'' For each
community, this process yields a balanced, labeled set of
controversial/non-controversial posts. Table~\ref{tab:basic_stats}
contains the number of posts/comments for each community after the
above filtration process, and the \pupv for the
controversial/non-controversial sets.

\begin{table}
  \small
  \centering
  \begin{tabular}{crrrrr}
\toprule
 & \# posts & \# cmnts & $\mu_{up}$ cont & $\mu_{up}$ noncont \\
\midrule
AM & 3.3K & 474K & 66\% & 90\% \\
AW & 3.0K & 417K & 67\% & 91\% \\
FT & 3.9K & 625K & 66\% & 91\% \\
LT & 1.6K & 208K & 68\% & 91\% \\
PF & 1.0K & 95K & 72\% & 92\% \\
RL & 2.2K & 221K & 68\% & 93\% \\
\bottomrule
\end{tabular}

  \caption{
        Dataset statistics: number of
    posts, number of comments, mean \pupv for the
    controversial and non-controversial classes.
  }
  \label{tab:basic_stats}
  \end{table}

\subsection{Quantitative validation of labels}
Reddit provides
a
sort-by-controversy function, and we wanted to ensure that our
controversy labeling method aligned with this
ranking.\footnote{
      This validation step rules out the
  possibility that \pupv is uncorrelated with Reddit's
  official definition of controversy.} We contacted Reddit itself, but
they were unable to provide details.
Hence, we scraped the 1K most controversial posts
according to Reddit (1K is the max that Reddit provides) for each community over the past year (as of October 2018). Next, we
sampled posts that \emph{did not} appear on Reddit's controversial list in
the year prior to October 2018
to create a
1:k ratio sample of Reddit-controversial posts and
non-Reddit-controversial posts for $k \in \{1,2,3\}$, $k=3$ being
the most difficult setting. Then, we applied the filtering/labeling
method described above, and measured how well our process matched
Reddit's ranking scheme, i.e., the ``controversy'' label applied by
our method matched the ``controversy'' label assigned by Reddit.

Our labeling method achieves high precision in identifying
controversial/non-controversial posts. While a large proportion of
posts are discarded, the labels assigned to surviving posts match those
assigned by Reddit with the following F-measures at $k=3$ (the results
for $k=1,2$ are higher):\footnote{There were communities that we did not consider
  because the correlation between our filter and Reddit's ranking was
  lower, e.g., \subreddit{PoliticalDiscussion}.}
\begin{center}
  {\footnotesize
  \begin{tabular}{l|cccccc}
    & AM & AW & FT & LT & PF & RL \\
    \midrule
        F-measure & 97 & 96 & 88 & 90 & 94 & 96 \\
  \end{tabular}
  }
\end{center}
In all cases, the precision for the non-controversial label is
perfect, i.e., our filtration method never labeled a
Reddit-controversial post as non-controversial. The precision of the
controversy label was also high, but imperfect; errors could be a
result of, e.g., Reddit's controversy ranking being limited to 1K
posts, or
using internal data, etc.

\subsection{Qualitative validation of labels} Figure~\ref{fig:examples}
gives examples of controversial and non-controversial posts from three
of the communities we consider, alongside the text of the first
comment made in response to those posts.

\mparagraph{Topical differences} A priori, we expect that the
topical content of posts may be related to how controversial they
become (see prior work in Fig.~\ref{fig:related_laid_out}).
  We ran LDA \cite{blei2003latent} with 10 topics on posts from each
community independently, and compared the differences in mean topic
frequency between controversial and non-controversial posts. We
observe community-specific patterns, e.g., in
\subreddit{relationships}, posts about family (top words in topic:
``family parents mom dad") are less controversial than those
associated with romantic relationships (top words: ``relationship,
love, time, life"); in \subreddit{AskWomen}, a gender topic (``women men
woman male'') tends to be associated with more controversy than an
advice-seeking topic (``im dont feel ive'')

\mparagraph{Wording differences} We utilize Monroe et al.'s
\shortcite{monroe2008fightin} algorithm for comparing language usage
in two bodies of text; the method places a
Dirichlet prior over n-grams (n=1,2,3)
and estimates Z-scores on the difference in
rate-usage between controversial and non-controversial
posts. 
This analysis reveals many community-specific patterns, e.g., phrases
associated with controversy include ``crossfit'' in
\subreddit{Fitness}, ``cheated on my'' in \subreddit{relationships},
etc. What's controversial in one community may be non-controversial in
another, e.g., ``my parents'' is associated with controversy in
\subreddit{personalfinance} (e.g., ``live with my parents'') but
strongly associated with lack of controversy in
\subreddit{relationships} (e.g., ``my parents got divorced''). We also
observe that some communities share commonalities in phrasing, e.g.,
``do you think'' is associated with controversy in both
\subreddit{AskMen} and \subreddit{AskWomen}, whereas ``what are some''
is associated with a lack of controversy in both.

\section{Early Discussion Threads}
\label{sec:comment-trees}

We now analyze comments posted in early
discussion threads for controversial vs.~non-controversial
posts.
In this section, we focus on comments posted
within one hour of the original submission, although
we consider a wider range of times in later experiments.

\mparagraph{Comment Text} We mirrored the n-gram analysis conducted in
the previous section, but, rather than the text of the original post, focused on the text of
comments. Many patterns persist, but the conversational framing
changes, e.g., ``I cheated'' in the {\em posts} of
\subreddit{relationships} is mirrored by ``you cheated'' in the
{\em comments}. Community differences
again appear: e.g., ``birth
control''
indicated controversy when it appears in
the comments for
\subreddit{relationships}, but not for \subreddit{AskWomen}.

\begin{figure}[t]
  \centering
  \begin{subfigure}{\linewidth}
  \begin{subfigure}{0.32\linewidth}
    \centering
    \includegraphics[width=.95\linewidth]{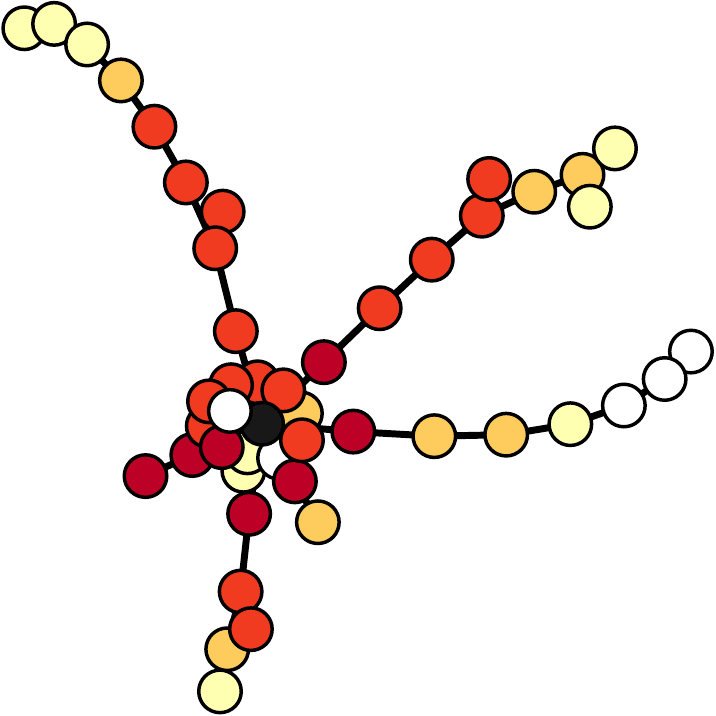}
  \end{subfigure}
  \begin{subfigure}{0.32\linewidth}
    \centering
    \includegraphics[width=.95\linewidth]{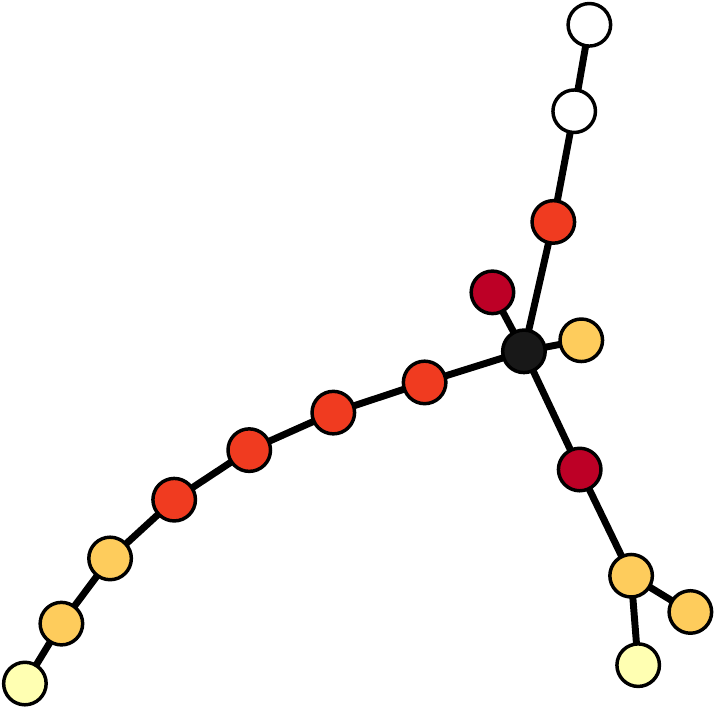}
  \end{subfigure}
  \begin{subfigure}{0.32\linewidth}
    \centering
    \includegraphics[width=.95\linewidth]{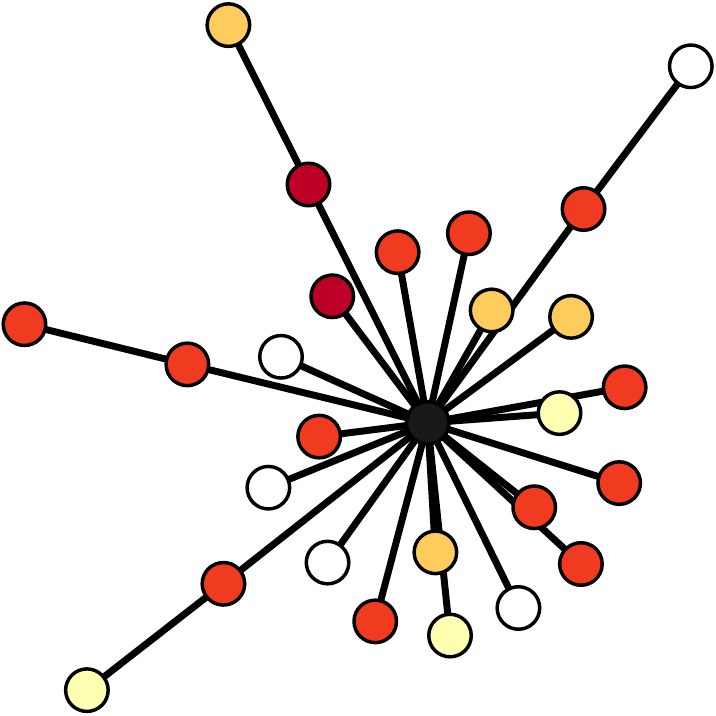}
  \end{subfigure}
  \caption{Discussions on controversial posts}
  \end{subfigure}

  \begin{subfigure}{\linewidth}
  \begin{subfigure}{0.32\linewidth}
    \centering
    \includegraphics[width=.95\linewidth]{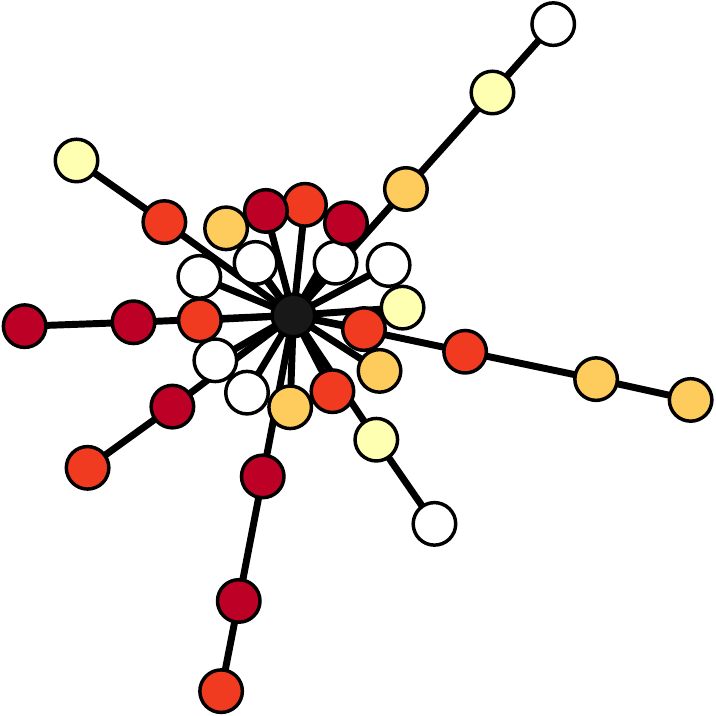}
  \end{subfigure}
  \begin{subfigure}{0.32\linewidth}
    \centering
    \includegraphics[width=.95\linewidth]{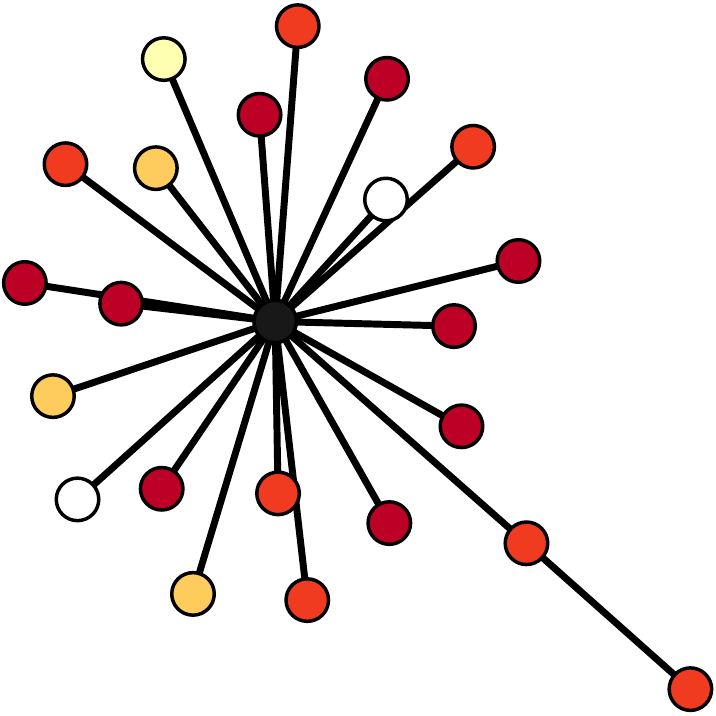}
  \end{subfigure}
  \begin{subfigure}{0.32\linewidth}
    \centering
    \includegraphics[width=.95\linewidth]{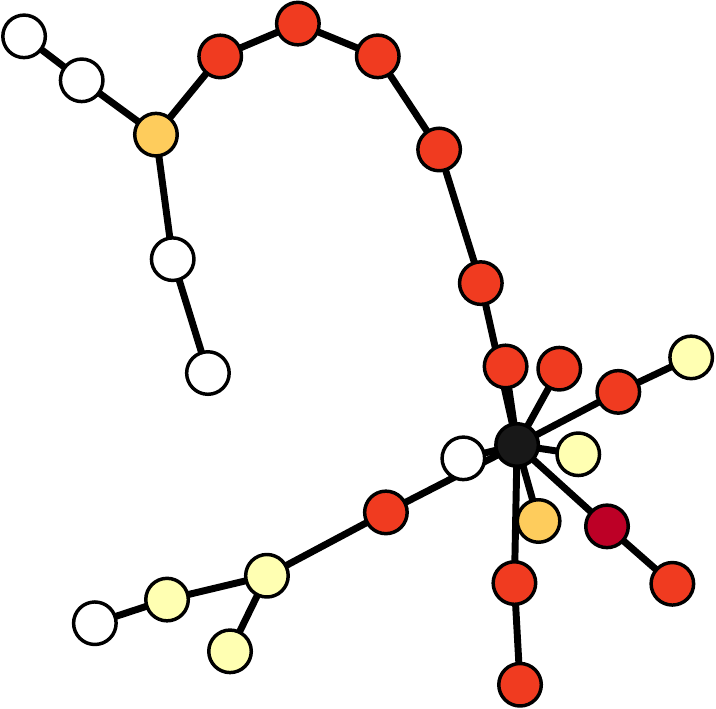}
  \end{subfigure}
  \caption{Discussions on non-controversial posts}
  \end{subfigure}
  \caption{Early conversation trees from \subreddit{AskMen}; nodes are
    comments and edges indicate reply structure. The original post is
    the black node, and as node colors lighten from red to yellow,
    comment timing increases from zero minutes to sixty minutes.}
  \label{fig:conv_trees}
\end{figure}

\mparagraph{Comment Tree Structure} While prior work in early
prediction mostly focuses on measuring rate of early responses, we
postulate that more expressive, structural features of conversation
trees may also carry predictive capacity.

Figure~\ref{fig:conv_trees} gives samples of conversation trees that
developed on Reddit posts within one hour of the original post
being made. There is significant diversity among tree size and
shape. To quantify these differences, we introduce two sets of
features: C-RATE features, which encode the rate of
commenting/number of comments;\footnote{
    Specifically: total number of comments, the logged time
  between OP and the first reply, and the average logged parent-child
  reply time over pairs of comments.}
and C-TREE features, which
encode structural aspects of discussion trees.\footnote{
      Specifically: max depth/total comment ratio,
  proportion of comments that were top-level (i.e., made in
  direct reply to the original post), average node depth, average
  branching factor, proportion of top-level comments replied
  to,
    Gini coefficient of replies to top-level
  comments (to measure how ``clustered'' the total discussion is), and
  Wiener Index of virality (which measures the average
  pairwise path-length between all nodes in the conversation tree
  \cite{wiener1947structural,goel2015structural}).  }
We then examine
whether or not tree features correlate with controversy after
controlling for popularity.

Using binary logistic regression, after controlling for C-RATE, C-TREE
features extracted from comments made within one hour of the original
post improve model fit in all cases except for
\subreddit{personalfinance} ($p<.05$, LL-Ratio test). We repeated the
experiment, but also controlled for \emph{eventual}
popularity\footnote{We added in the logged number of eventual
  comments, and also whether or not the post received an above-median number of
  comments.} in addition to C-RATE, and observed the same result. This
provides evidence that \emph{structural} features of conversation
trees are predictive, though \emph{which} tree feature is most
important according to these experiments is community-specific. For
example, for the models without eventual popularity information, the
C-TREE feature with largest coefficient in \subreddit{AskWomen} and
\subreddit{AskMen} was the max-depth ratio,
but it was the Wiener index
in \subreddit{Fitness}.

\section{Early Prediction of Controversy}

We shift our focus to the task of predicting controversy on Reddit.
In general, tools that predict controversy are most useful if they
only require information available at the time of submission or
as soon as possible thereafter.
We note
that while the causal relationship between vote totals and comment
threads is not entirely clear (e.g., perhaps the comment threads cause
more up/down votes on the post), predicting the ultimate
\emph{outcome} of posts is still useful for community moderators.

\mparagraph{Experimental protocols}
All classifiers are binary
(i.e., controversial vs.~non-controversial) and, because the classes
are in 50/50 balance, we compare algorithms according to their
accuracy. Experiments are conducted as 15-fold cross validation with
random 60/20/20 train/dev/test splits, where the splits are drawn to
preserve the 50/50 label distribution. For non-neural, feature-based
classifiers, we use linear models.\footnote{We cross-validate
  regularization strength 10\^{}(-100,-5,-4,-3,-2,-1,0,1), model type
  (SVM vs. Logistic L1 vs. Logistic L2 vs. Logistic L1/L2), and
  whether or not to apply feature standardization for each feature set
  and cross-validation split separately. These are trained using
  \texttt{lightning}
  (\url{http://contrib.scikit-learn.org/lightning/}).} For BiLSTM
models,\footnote{We optimize using Adam \cite{kingma2014adam} with
  LR=.001 for 20 epochs, apply dropout with $p=.2$, select the model
  checkpoint that performs best over the validation set, and
  cross-validate the model's dimension (128 vs. 256) and the number of
  layers (1 vs. 2) separately for each cross-validation split.} we use
Tensorflow \cite{tensorflow2015-whitepaper}. Whenever a feature is
ill-defined (e.g., if it is a comment text feature, but there are no
comments at time $t$) the column mean of the training set for each
cross-validation split is substituted. Similarly, if a comment's body
is deleted, it is ignored by text processing algorithms. We perform
both Wilcoxon signed-rank tests \cite{demvsar2006statistical} and
two-sided corrected resampled t-tests \cite{nadeau2000inference} to
estimate statistical significance, taking the maximum of the two
resulting p-values to err on the conservative side and reduce the
chance of Type I error.

\subsection{Comparing text models}

The goal of this section is to compare text-only models for
classifying controversial vs. non-controversial posts. Algorithms are
given access to the full post titles and bodies, unless stated otherwise.

\mparagraph{HAND} We consider a number of hand-designed features
related to the textual content of posts inspired by
\newcite{tan2016winning}.\footnote{Specifically:
  for the title and text body separately, length, type-token ratio,
rate of first-person pronouns, rate of second-person pronouns, rate of
question-marks, rate of capitalization, and Vader sentiment
\cite{gilbert2014vader}. Combining
the post title and post body: number of links, number of Reddit links,
number of imgur links,
number of sentences, Flesch-Kincaid readability
score, rate of italics, rate of boldface, presence of a list, and the rate
of word use from 25 Empath wordlists
\cite{fast2016empath}, which include various categories,
such as politeness, swearing, sadness, etc.}

\mparagraph{TFIDF} We encode posts according to tfidf feature
vectors. Words are included in the vocabulary if they appear more than
5 times in the corresponding cross-validation split.

\mparagraph{W2V} We consider a mean, 300D word2vec
\cite{mikolov2013distributed} embedding representation, computed from
a GoogleNews corpus.

\mparagraph{ARORA} A slight modification of W2V, proposed by
\newcite{arora2016simple}, serves as a ``tough to beat'' baseline
for sentence representations.

\mparagraph{LSTM} We train a Bi-LSTM \cite{graves2005framewise} over
the first 128 tokens of titles + post text, followed by a mean pooling
layer, and then a logistic regression layer. The LSTM's embedding
layer is initialized with the same word2vec embeddings used in W2V.
Markdown formatting artifacts are discarded.

\mparagraph{BERT-LSTM} Recently, features extracted from fixed,
pretrained, neural language models have resulted in high performance
on a range of language tasks. Following the recommendations of \S5.4
of \newcite{devlin2018bert}, we consider representing posts by
extracting BERT-Large embeddings computed for the first 128 tokens of
titles + post text; we average the final 4 layers of the 24-layer,
pretrained Transformer-decoder network
\cite{vaswani2017attention}. These token-specific vectors are then
passed to a Bi-LSTM, a mean pooling layer, and a logistic
classification layer. We keep markdown formatting artifacts because
BERT's token vocabulary are WordPiece subtokens \cite{wu2016google},
which are able to incorporate arbitrary punctuation without
modification.

\mparagraph{BERT-MP} Instead of training a Bi-LSTM over BERT features,
we mean pool over the first 128 tokens, apply L2 normalization to the
resulting representations, reduce to 100 dimensions using
PCA,\footnote{Values of 50 and 150 both work well, too.} and train a
linear classifier on top.

\mparagraph{BERT-MP-512} The same as BERT-MP, except the algorithm is
given access to 512 tokens (the maximum allowed by BERT-Large) instead
of 128.

\begin{table}
  \centering
  \scriptsize
  \setulcolor{lightgray}
\begin{tabular}{l|cccccc}
\toprule
 & AM & AW & FT & LT & PF & RL\\
\midrule
HAND & 55.4 & 52.2 & 61.9 & 59.7 & 54.5 & 60.8\\
TFIDF & 57.4 & 60.1 & 63.3 & 59.1 & 58.7 & 65.4\\
ARORA & 58.6 & 62.0 & 60.5 & 59.4 & 57.2 & 62.1\\
W2V & 60.7 & 62.1 & 63.1 & 61.4 & 59.9 & 64.3\\
\midrule
LSTM & 58.9 & 58.2 & 63.6 & 61.5 & 60.0 & 63.1\\
BERT-LSTM & \textbf{64.5} & \textbf{65.1} & \textbf{66.2} & \ul{65.0} & \ul{65.1} & \ul{67.8}\\
BERT-MP & \ul{63.4} & \ul{64.0} & \ul{64.4} & \ul{65.7} & \ul{64.1} & \ul{67.0}\\
BERT-MP-512 & \ul{63.9} & \ul{64.0} & \ul{64.7} & \textbf{65.8} & \textbf{65.6} & \ul{67.7}\\
\midrule
HAND+W2V & 61.3 & 62.3 & \ul{64.9} & \ul{63.2} & 60.0 & \ul{66.3}\\
HAND+BERTMP512 & \ul{63.6} & \ul{63.5} & \ul{64.9} & \ul{64.1} & \ul{64.4} & \textbf{68.0}\\
\bottomrule
\end{tabular}
\setulcolor{black}

  \caption{Average accuracy for each post-time, text-only predictor for each dataset,
    averaged over 15 cross-validation splits; standard errors are
    $\pm.6$, on average (and never exceed $\pm1.03$). Bold is
    best in column; underlined are statistically
    indistinguishable from best in column ($p<.01$)}
  \label{tab:text_results}
\end{table}

\mparagraphnp{Results:} Table~\ref{tab:text_results} gives the
performance of each text classifier for each community. In general,
the best performing models are based on the BERT features, though
HAND+W2V performs well, too. However, no performance gain is achieved
when adding hand designed features to BERT. This may be because BERT's
subtokenization scheme incorporates punctuation, link urls, etc.,
which are similar to the features captured by HAND. Adding an LSTM
over BERT features is comparable to mean pooling over the sequence;
similarly, considering 128 tokens vs.~512 tokens results in comparable
performance. Based on the results of this experiment, we adopt
BERT-MP-512 to represent text in experiments for the rest of this
work.

\begin{figure*}
  \centering
  \begin{subfigure}{.3\linewidth}
    \centering
    \includegraphics[width=.85\linewidth]{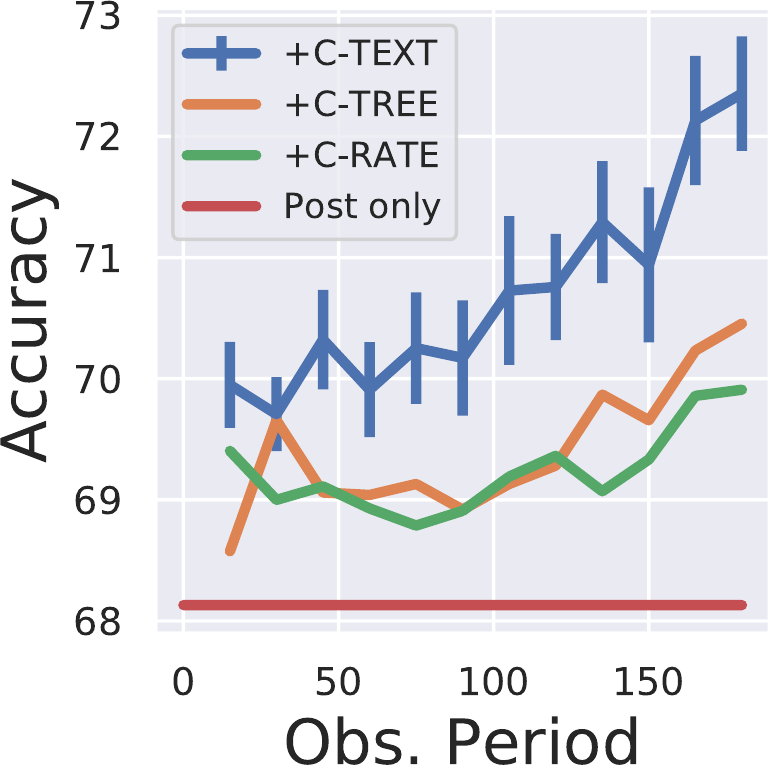}
    \caption{\subreddit{AskMen} ($t_s=15$)}\label{fig:AskMen_time}
  \end{subfigure}
  \hfill
  \begin{subfigure}{.3\linewidth}
    \centering
    \includegraphics[width=.85\linewidth]{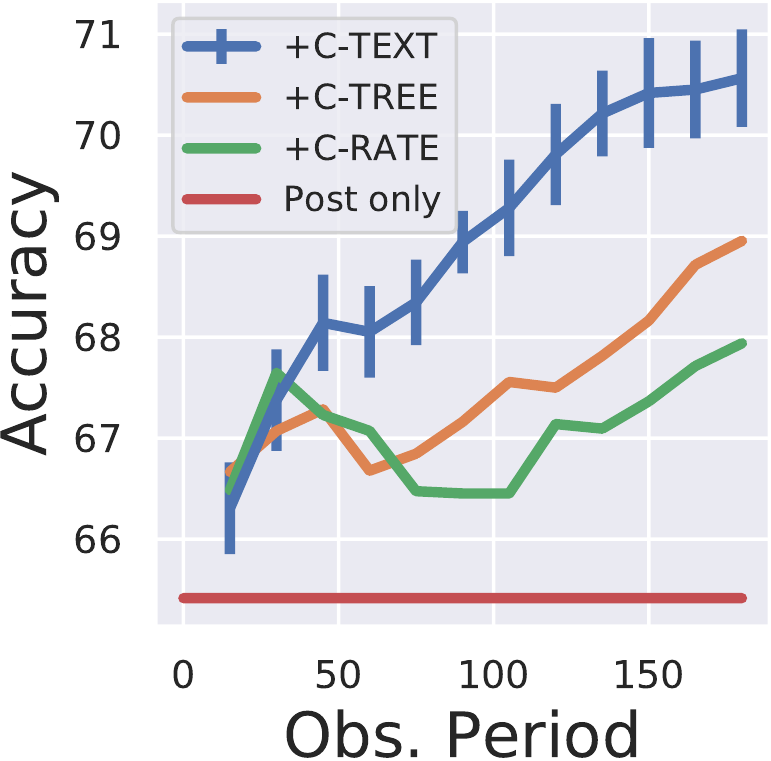}
    \caption{\subreddit{AskWomen} ($t_s=45$)}\label{fig:AskWomen_time}
  \end{subfigure}
  \hfill
  \begin{subfigure}{.3\linewidth}
    \centering
    \includegraphics[width=.85\linewidth]{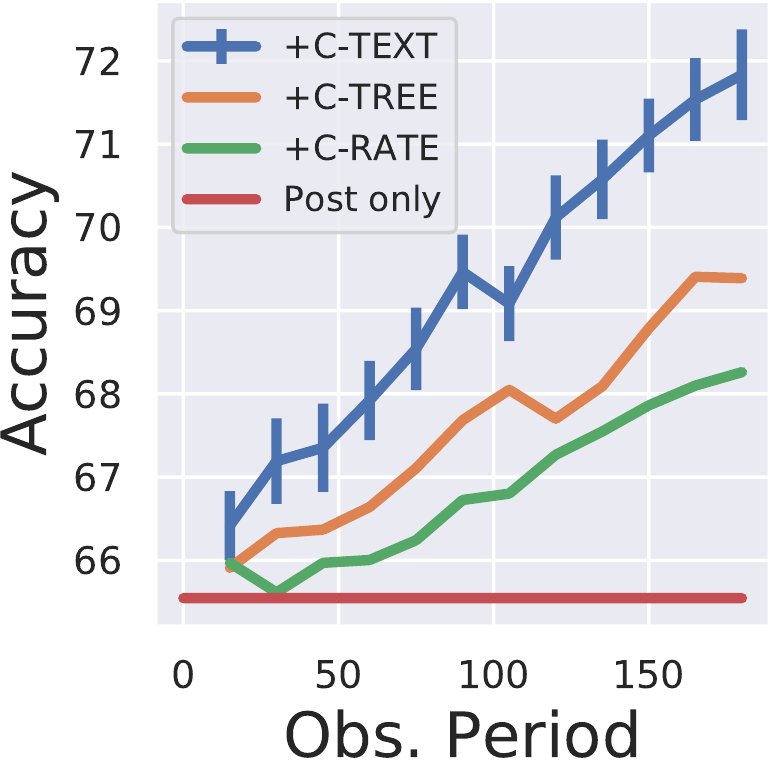}
    \caption{\subreddit{Fitness ($t_s=60$)}}\label{fig:Fitness_time}
  \end{subfigure}

  \bigskip
  \begin{subfigure}{.3\linewidth}
    \centering
    \includegraphics[width=.85\linewidth]{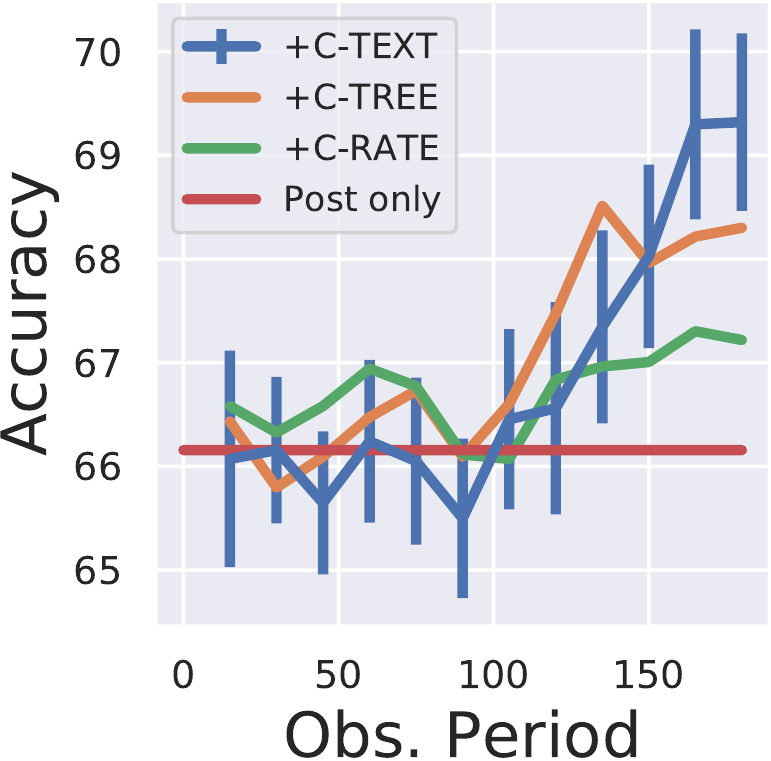}
    \caption{\subreddit{LifeProTips} ($t=165$)}\label{fig:LifeProTips_time}
  \end{subfigure}
  \hfill
  \begin{subfigure}{.3\linewidth}
    \centering
    \includegraphics[width=.85\linewidth]{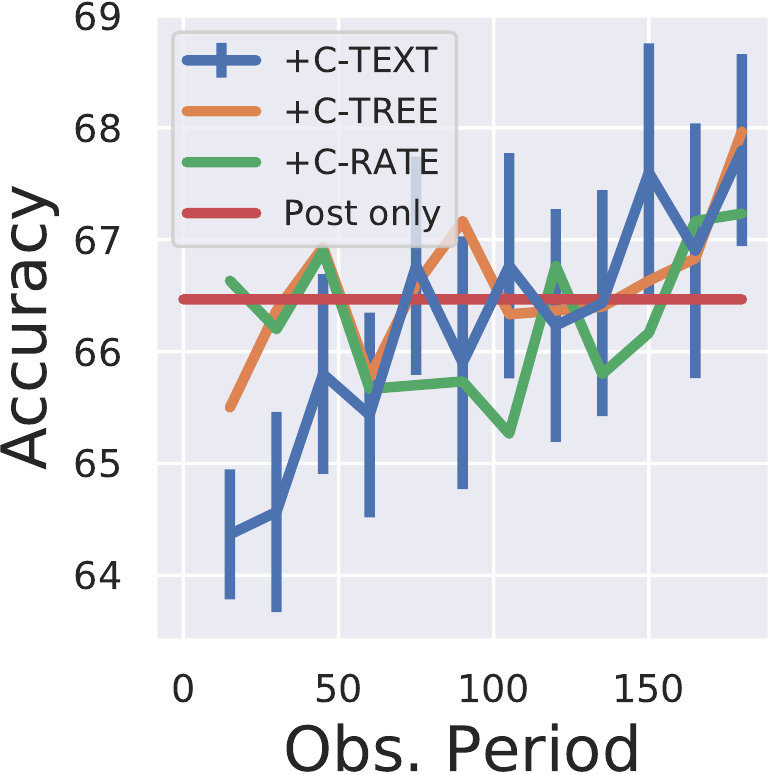}
    \caption{\subreddit{personalfinance} (N/A)}\label{fig:personalfinance_time}
  \end{subfigure}
  \hfill
  \begin{subfigure}{.3\linewidth}
    \centering
    \includegraphics[width=.85\linewidth]{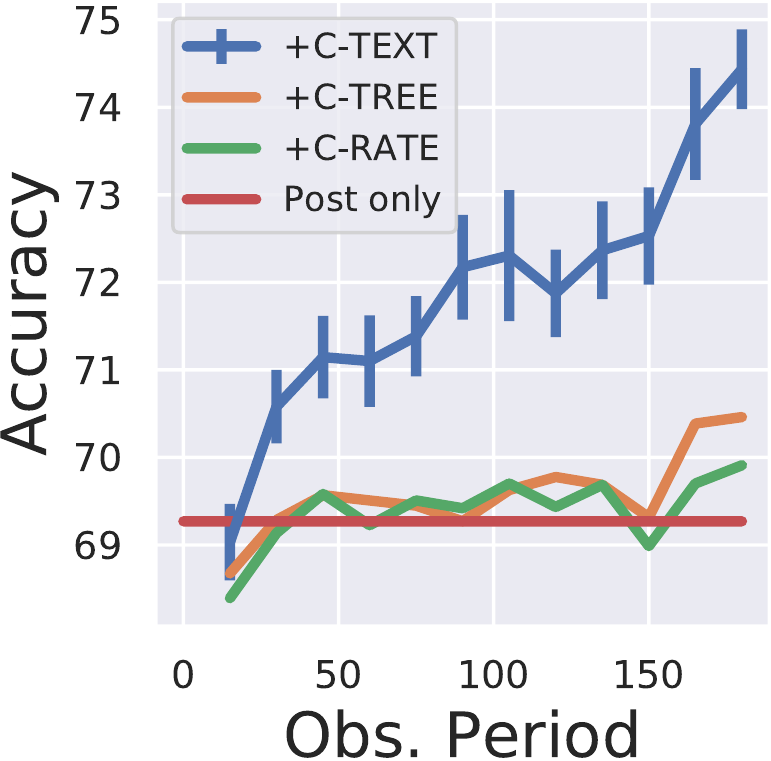}
    \caption{\subreddit{relationships} ($t_s=45$)}\label{fig:relationships_time}
  \end{subfigure}

  \caption{Classifier accuracy for increasing periods of observation;
    the ``+'' in the legend indicates that a feature set is combined
    with the feature sets below. $t_s$, the time the full feature set first achieves statistical
    significance over the post-time only baseline, is given for each
    community (if significance is achieved).
    }
  \label{fig:time_sweep}
\end{figure*}

\subsection{Post-time Metadata}

\begin{table}
  \scriptsize
  \centering
  \begin{tabular}{l|cccccc}
\toprule
 & AM & AW & FT & LT & PF & RL\\
\midrule
TEXT & 63.9 & 64.0 & 64.7 & 65.8 & 65.6 & 67.7\\
+TIME & 68.1 & 65.4 & 65.5 & 66.2 & 66.5 & 69.3\\
+AUTHOR & 68.2 & 65.3 & 65.7 & 66.0 & 66.4 & 69.3\\
\bottomrule
\end{tabular}

  \caption{Post-time only results: the effect of incorporating timing
    and author identity features.}
  \label{tab:text_meta_results}
\end{table}

Many non-content factors can influence community reception of posts,
e.g., \newcite{hessel2017cats} find that \emph{when} a post is made on
Reddit can significantly influence its eventual popularity.

\mparagraph{TIME} These features encode when a post was created. These
include indicator variables for year, month, day-of-week, and
hour-of-day.

\mparagraph{AUTHOR} We add an indicator variable for each user that
appears at least 3 times in the training set, encoding the hypothesis
that some users may simply have a greater propensity to post
controversial content.

The results of incorporating the metadata features on top of TEXT are
given in Table~\ref{tab:text_meta_results}. While incorporating TIME
features on top of TEXT results in consistent improvements across all
communities,\jack{These arent statistically significant, outside of
  askmen} incorporating author features on top of TIME+TEXT does
not. We adopt our highest performing models, TEXT+TIME, as a strong
post-time baseline.

\subsection{Early discussion features}

\begin{figure}
  \begin{minipage}{0.475\linewidth}
    \centering
    \includegraphics[width=.8\linewidth]{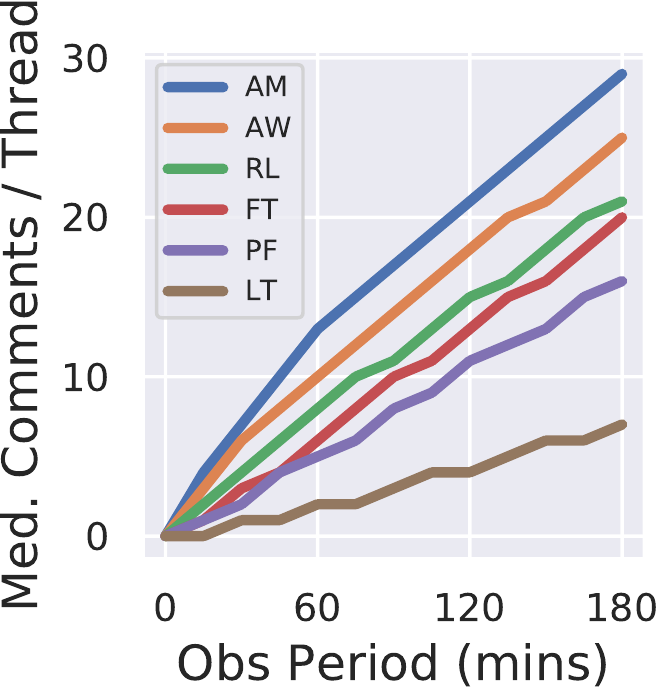}
    \captionof{figure}{Observation period versus median number of
      comments available.}
    \label{fig:discard_prop}
  \end{minipage}  \hfill
  \begin{minipage}{0.475\linewidth}
    \centering
    \includegraphics[width=.8\linewidth]{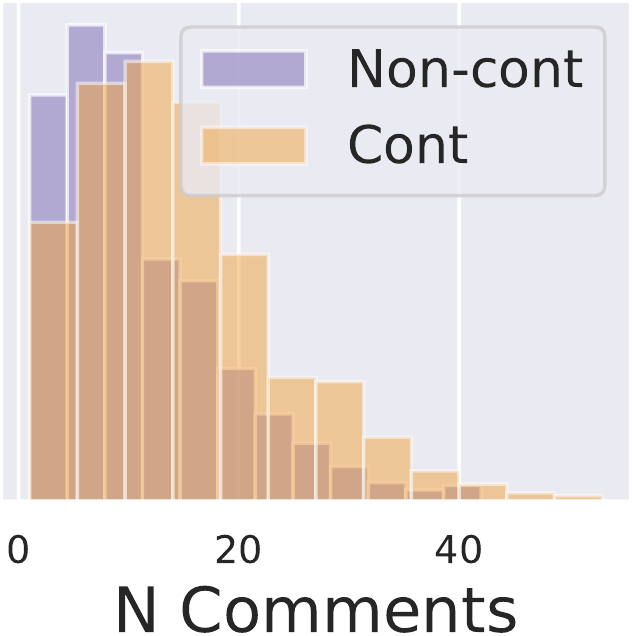}
    \captionof{figure}{Histogram of the number of comments available per thread at $t=60$ minutes in \subreddit{AskWomen}.}
    \label{fig:n_comments_60_histogram}
  \end{minipage}
\end{figure}

\mparagraph{Basic statistics of early comments} We augment the
post-time features with early-discussion feature sets by giving our
algorithms access to comments from increasing observation
periods. Specifically, we train linear classifiers by combining our
best post-time feature set (TEXT+TIME) with features derived from
comment trees available after $t$ minutes, and sweep $t$ from $t=15$
to $t=180$ minutes in 15 minute intervals.

Figure~\ref{fig:discard_prop} plots the median number of comments
available per thread at different $t$ values for each community. The
amount of data available for the early-prediction algorithms to
consider varies significantly, e.g., while \subreddit{AskMen} threads
have a median 10 comments available at 45 minutes,
\subreddit{LifeProTips} posts do not reach that threshold even after 3
hours, and we thus expect that it will be a harder setting for early
prediction. We see, too, that even our maximal 3 hour window is
still early in a post's lifecycle, i.e., posts tend to receive significant attention
afterwards:  only 15\% (LT) to 32\% (AW) of all eventual comments are
available per thread at this time, on average.
Figure~\ref{fig:n_comments_60_histogram} gives the distribution of the
number of comments available for controversial/non-controversial posts
on \subreddit{AskWomen} at $t=60$ minutes. As with the other
communities we consider, the distribution of number of available posts
is not overly-skewed, i.e., most posts in our set (we filtered out
posts with less than 30 comments) get at least \emph{some} early
comments.

We explore a number of feature sets based on early comment
trees (comment feature sets are prefixed with ``C-''):

\mparagraph{C-RATE and C-TREE}
We described these in \S\ref{sec:comment-trees}.

\mparagraph{C-TEXT} For each comment available at a given observation
period, we extract the BERT-MP-512 embedding. Then, for each
conversation thread, we take a simple mean over all comment
representations. While we tried several more expressive means of
encoding the text of posts in comment trees, this simple method proved
surprisingly effective.\footnote{We do not claim that this is the
  \emph{best} way to represent text in comment trees. However, this
  simple method produces performance improvements over strong
  post-time baselines; exploring better models is a promising avenue
  for future work.}

\mparagraph{Sweeping over time} Figure~\ref{fig:time_sweep} gives the
performance of the post-time baseline combined with comment features
while sweeping $t$ from 15 to 180 minutes. For five of the six communities we
consider, the performance of the comment feature classifier
significantly ($p<.05$) exceeds the performance of the post-time
baseline in less than three hours of observation, e.g., in the case of
\subreddit{AskMen} and \subreddit{AskWomen}, significance is achieved
within 15 and 45 minutes, respectively.

In general, C-RATE improves only slightly over post only, even though
rate features have proven useful in predicting popularity in prior
work \cite{he2014predicting}. While adding C-TREE also improves
performance, comment textual content is the biggest source of
predictive gain. These results demonstrate i) that incorporating a
variety of early conversation features, e.g., structural features of
trees, can improve performance of controversy prediction
over strong post-time baselines, and ii) the text content of comments
contains significant complementary information to post text.

\mparagraph{Controversy prediction $\neq$ popularity prediction} We
return to a null hypothesis introduced in \S\ref{sec:dataset}: that
the controversy prediction models we consider here are merely learning
the same patterns that a popularity prediction algorithm would
learn. We train popularity prediction algorithms, and then attempt to
use them at test-time to predict controversy; under the null
hypothesis, we would expect little to no performance degradation when
training on these alternate labels.

We 1) train binary popularity predictors using post text/time +
comment rate/tree/text features available at $t=180$,\footnote{We
  predict whether or not a post eventually receives an above-median
  number of comments. We force the popularity predictors to predict
  50/50 at test time, which improves their performance.} and use them
to predict controversy at test-time; and 2) consider an oracle that
predicts the true popularity label at test-time; this oracle is
\emph{quite} strong, as prior work suggests that perfectly predicting
popularity is impossible \cite{salganik2006experimental}.
\begin{center}
  {\scriptsize
  \begin{tabular}{l|cccccc}
    & AM & AW & FT & LT & PF & RL \\
    \midrule
        {\small Pop Pred} & 53.9 & 55.2 & 60.1 & 54.2 & 52.9 & 52.8 \\
        {\small Pop Oracle} & 65.8 & 67.0 & 70.3 & 68.1 & 64.0 & 63.3 \\
  \end{tabular}
  }
\end{center}

In all cases, the best popularity predictor does not achieve
performance comparable to even the post-only baseline. For 3 of 6
communities, even the popularity oracle does not beat post time
baseline, and in all cases, the mean performance of the controversy
predictor exceeds the oracle by $t=180$. Thus, in our setting,
controversy predictors and popularity predictors learn disjoint
patterns.

\subsubsection{Domain Transfer} \label{ssub:domain_transfer}

We conduct experiments where we train
models on one subreddit and test them on another. For these
experiments, we discard all posting time features, and compare
C-(TEXT+TREE+RATE) to C-(TREE+RATE); the goal is to empirically examine the
hypothesis in \S\ref{sec:intro}: that controversial text is
community-specific.

To measure performance differences in the domain transfer setting, we
compute the percentage accuracy drop relative to a constant prediction
baseline when switching the training subreddit from the matching
subreddit to a different one. For example, at $t=60$, we observe
that raw accuracy drops from $65.6 \rightarrow 55.8$ when training on
\subreddit{AskWomen} and testing on \subreddit{AskMen} when
considering text, rate, and tree features together; given that the
constant prediction baseline achieves 50\% accuracy, we compute the
percent drop in accuracy as: $(55.8-50)/(65.6-50)-1 = -63\%$.

\begin{figure}
  \centering
  \begin{subfigure}{.45\linewidth}
    \captionsetup{justification=centering}
    \centering
    \includegraphics[width=.95\linewidth]{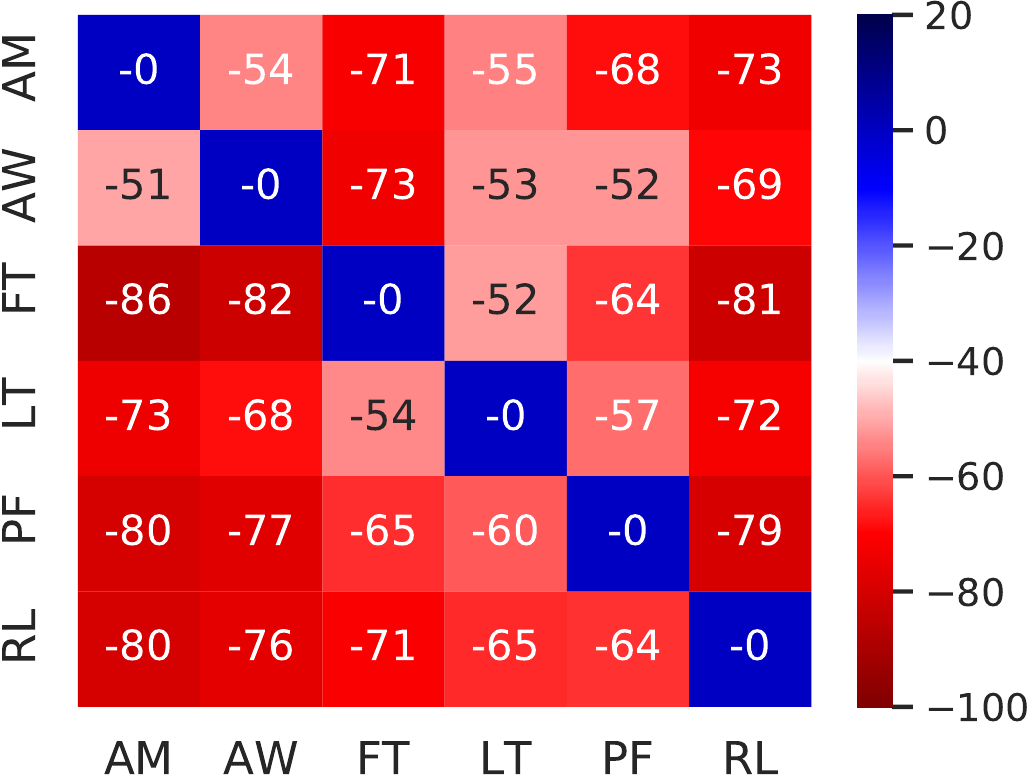}
    \caption{{\scriptsize TEXT+RATE+TREE\\$t=180$}}
    \label{fig:transfer_text_rate_and_tree_northwest}
  \end{subfigure}  \begin{subfigure}{.45\linewidth}
    \captionsetup{justification=centering}
    \centering
    \includegraphics[width=.95\linewidth]{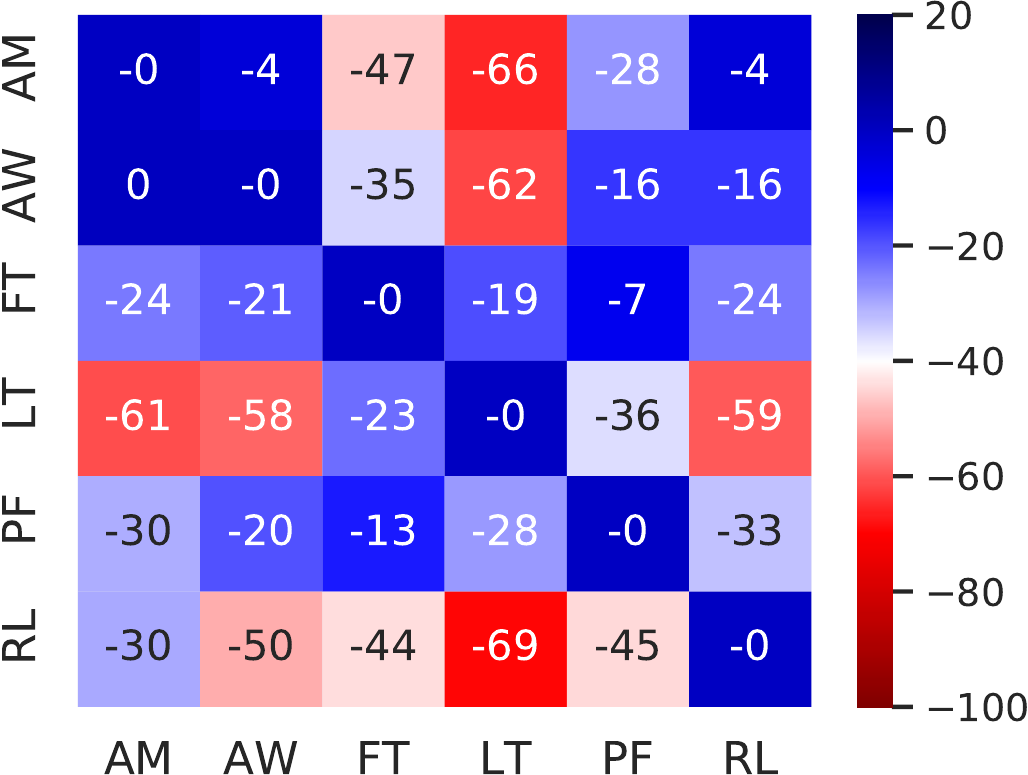}
    \caption{{\scriptsize RATE+TREE\\$t=180$}}
    \label{fig:transfer_rate_and_tree_northeast}
  \end{subfigure}
  \\
  \hrule
  \vspace{.15cm}
  \begin{subfigure}{.45\linewidth}
    \captionsetup{justification=centering}
    \centering
    \includegraphics[width=.95\linewidth]{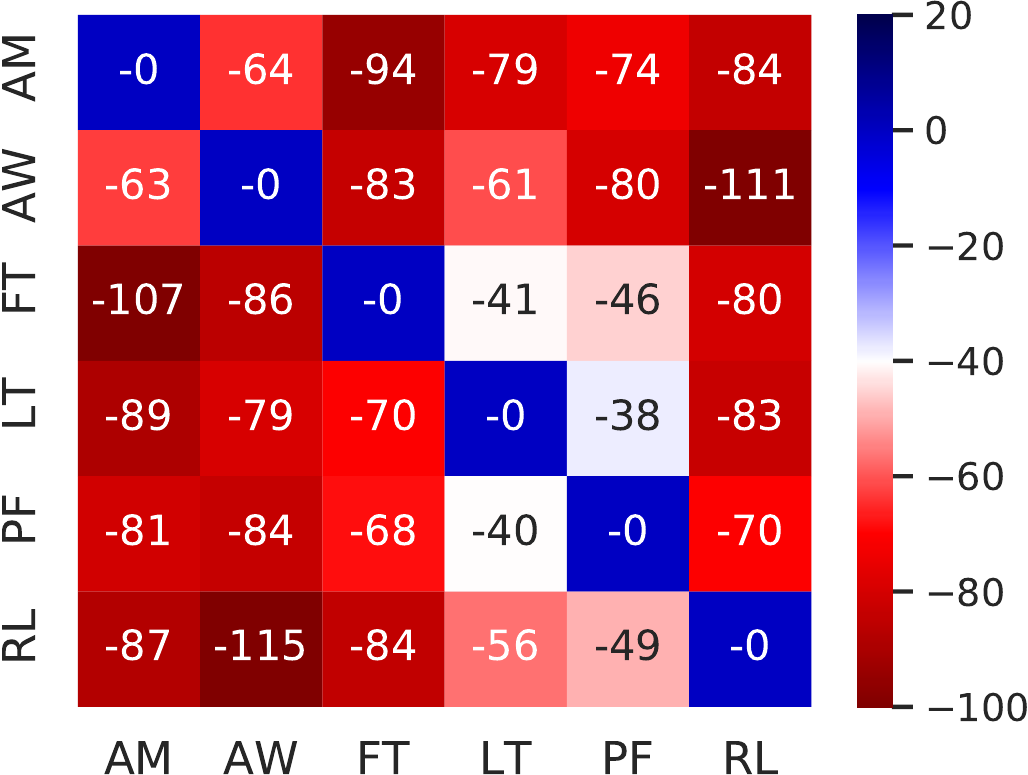}
    \caption{{\scriptsize TEXT+RATE+TREE\\$t=60$}}
    \label{fig:transfer_text_rate_and_tree_southwest}
  \end{subfigure}  \begin{subfigure}{.45\linewidth}
    \captionsetup{justification=centering}
    \centering
    \includegraphics[width=.95\linewidth]{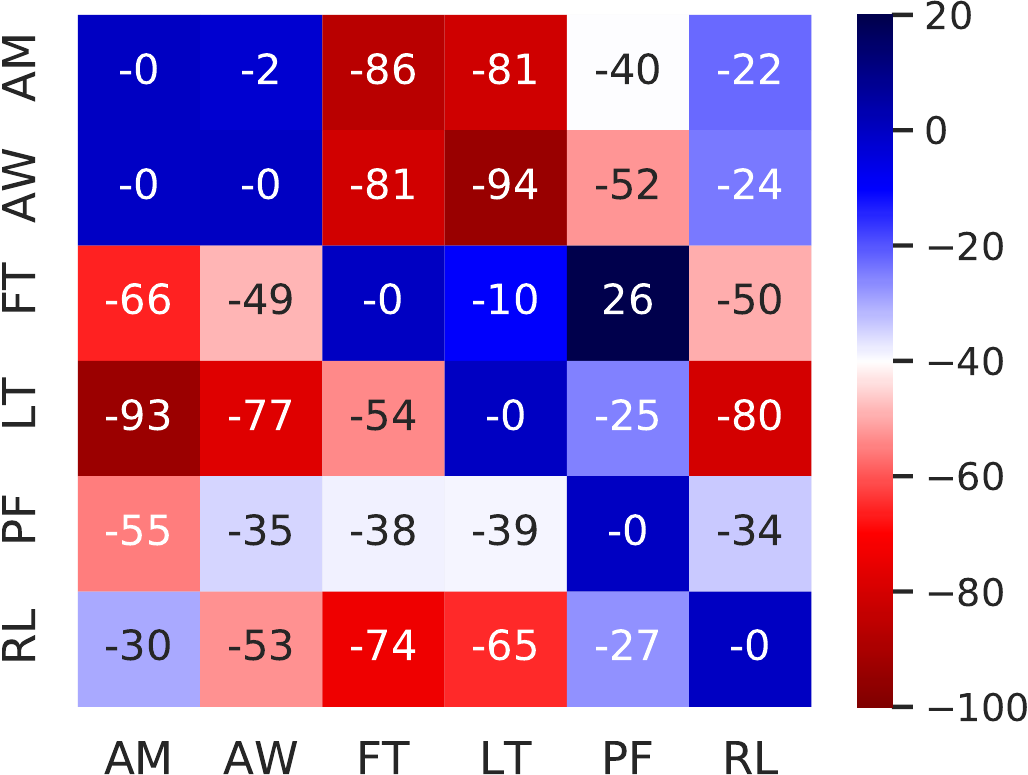}
    \caption{{\scriptsize RATE+TREE\\$t=60$}}
    \label{fig:transfer_rate_and_tree_southeast}
  \end{subfigure}
  \caption{Average cross-validated performance degradation for
    transfer learning setting at $t=180$ and $t=60$; the y-axis is the
    training subreddit and the x-axis is testing. For a fixed test
    subreddit, each column gives the percent accuracy drop when
    switching from the matching training set to a domain transfer
    setting. In general, while incorporating comment text features
    results in higher accuracy overall, comment rate + tree features
    transfer between communities with less performance degradation.}
    \label{fig:transfer_learning}
\end{figure}

The results of this experiment (Figure~\ref{fig:transfer_learning})
suggest that while text features are quite strong in-domain, they are
brittle and community specific. Conversely, while rate and structural
comment tree features do not carry as much in-domain predictive
capacity on their own, they generally transfer better between
communities, e.g., for RATE+TREE, there is very little performance
drop-off when training/testing on
\subreddit{AskMen}/\subreddit{AskWomen} (this holds for all timing
cutoffs we considered). Similarly, in the case of training on
\subreddit{Fitness} and testing on \subreddit{PersonalFinance}, we
sometimes observe a performance \emph{increase} when switching domains
(e.g., at $t=60$); we suspect that this could be an effect of dataset
size, as our \subreddit{Fitness} dataset has the most posts of any
subreddit we consider, and \subreddit{PersonalFinance} has the least.

\section{Conclusion}
\label{sec:conclusion}

We demonstrated that early discussion features are predictive of
eventual controversiality in several reddit communities. This finding
was dependent upon considering an expressive feature set of early
discussions; to our knowledge, this type of feature set (consisting of
text, trees, etc.) hadn't been thoroughly explored in prior early
prediction work.

One promising avenue for future work is to examine higher-quality
textual representations for conversation trees. While our mean-pooling
method did produce high performance, the resulting classifiers do not
transfer between domains effectively. Developing a more expressive
algorithm (e.g., one that incorporates reply-structure relationships)
could boost predictive performance, and enable textual features to be
less brittle.

\newcommand{\firstname}[2]{#1}
{\small
\paragraph*{Acknowledgments}
We thank
\firstname{Cristian}{C.} Danescu-Niculescu-Mizil,
Justine Zhang,
Vlad Niculae,
Jon Kleinberg,
and the anonymous reviewers for their helpful feedback.
We additionally thank
NVidia Corporation for the GPUs used in this study.
This work was supported in part by NSF grant SES-1741441.  Any opinions, findings, and conclusions or recommendations expressed in this material are those of the authors and do not necessarily
reflect the views of the sponsors. \par
}
\bibliography{refs}
\bibliographystyle{acl_natbib}

\end{document}